\documentclass[sn-mathphys]{sn-jnl}

\usepackage[normalem]{ulem}

\jyear{2023}%

\theoremstyle{thmstyleone}%
\theoremstyle{thmstyletwo}%

\theoremstyle{thmstylethree}%
\newcommand{\eg}{{\it e.g.}}

\begin{document}

\title[Robust Camera Pose Estimation for Endoscopic Videos]{Learning How To Robustly Estimate Camera Pose in Endoscopic Videos}


\author*[1]{\fnm{Michel} \sur{Hayoz}}\email{michel.hayoz@unibe.ch}
\author[1]{\fnm{Christopher} \sur{Hahne}}
\author[1]{\fnm{Mathias} \sur{Gallardo}}
\author[2]{\fnm{Daniel} \sur{Candinas}}
\author[3]{\fnm{Thomas} \sur{Kurmann}}
\author[3]{\fnm{Maximilian} \sur{Allan}}
\author[1]{\fnm{Raphael} \sur{Sznitman}}
\affil[1]{\orgdiv{ARTORG Center}, \orgname{University of Bern}, \country{Switzerland}}
\affil[2]{\orgdiv{Dept. of Visceral Surgery and Medicine}, \orgname{Inselspital}, \country{Switzerland}}
\affil[3]{\orgdiv{ Applied Research}, \orgname{Intuitive Surgical}, \country{USA}}




\abstract{
\textbf{Purpose:} Surgical scene understanding plays a critical role in the technology stack of tomorrow's intervention-assisting systems in endoscopic surgeries.
For this, tracking the endoscope pose is a key component, but remains challenging due to illumination conditions, deforming tissues and the breathing motion of organs.
\textbf{Method:} We propose a solution for stereo endoscopes that estimates depth and optical flow to minimize two geometric losses for camera pose estimation.
Most importantly, we introduce two learned adaptive per-pixel weight mappings that balance contributions according to the input image content.
To do so, we train a Deep Declarative Network to take advantage of the expressiveness of deep-learning and the robustness of a novel geometric-based optimization approach.
We validate our approach on the publicly available SCARED dataset and introduce a new in-vivo dataset, {\it StereoMIS}, which includes a wider spectrum of typically observed surgical settings.
\textbf{Results:} Our method outperforms state-of-the-art methods on average and more importantly, in difficult scenarios where tissue deformations and breathing motion are visible. We observed that our proposed weight mappings attenuate the contribution of pixels on ambiguous regions of the images, such as deforming tissues.
\textbf{Conclusion:} We demonstrate the effectiveness of our solution to robustly estimate the camera pose in challenging endoscopic surgical scenes. Our contributions can be used to improve related tasks like simultaneous localization and mapping (SLAM) or 3D reconstruction, therefore advancing surgical scene understanding in minimally-invasive surgery.  
}


\maketitle

\section{Introduction}
\label{sec:intro}

Camera pose estimation is a well established computer vision problem at the core of numerous applications of medical robotic systems for minimally invasive surgery (MIS). With a variety of methods proposed in recent years, most approaches have focused on Simultaneous Localization and Mapping (SLAM) and Visual Odometry (VO) frameworks to solve the pose estimation problem.
Well established techniques such as ORB-SLAM2~\cite{Mur-Artal2017} and ElasticFusion~\cite{Whelan2016} have shown great promise in rigid scenes.
More recently, non-rigid cases in MIS using monocular~\cite{Lamarca2019, Gomez-Rodriguez2021, Liu2022} and stereoscopic cameras~\cite{Song2018, Zhou2021, Ruofeng2022} have also been studied. 
Yet to this day, pose estimation in typical MIS settings remains particularly difficult due to deformations caused by instruments and breathing, self or instrument-based occlusions, textureless surfaces and tissue specularities.

In this work, we tackle the problem of pose estimation in such difficult cases when using a stereo endoscopic camera system. 

This allows depth estimation to be computed from parallax, which has been shown to improve robustness of SLAM methods~\cite{Mur-Artal2017}. In contrast to~\cite{Song2018, Zhou2021} which assume the tissue is smooth and locally rigid, respectively, we avoid making assumptions on the tissue deformation and topology. Instead, we propose a dense stereo VO framework that handles tissue deformations and the complexity of surgical scenes. To do this, our approach leverages geometric pose optimization by infering where to look at in the scene. At its core, our method uses a Deep Declarative Network (DDN)~\cite{Gould2021} to enable back-propagation of gradients through the pose optimization. 

More specifically, we propose to integrate two adaptive weight maps with the role of balancing the contribution of two geometrical losses and the contribution of each pixel towards each of these losses. We learn these adaptive weight maps using a DDN with the goal of solving the pose estimation problem, inspired by the recent DiffPoseNet~\cite{Parameshwara} approach.
Similarly to theirs, our method exploits the expressiveness of neural networks while leveraging robustness from the geometric optimization approach. This allows our method to adapt the contribution of the image region depending on the image content, for each loss but also between the two losses. 
We thoroughly evaluate our approach by characterizing its performance in comparison to the state-of-the-art on various practical scenarios: rigid scenes, breathing, scanning and deforming sequences. This validation is performed on two different datasets and we show that our method allows for more precise pose estimation in a wide range of cases.

\section{Method}

\begin{figure}[!b]
    \centering
    \includegraphics[width=\linewidth]{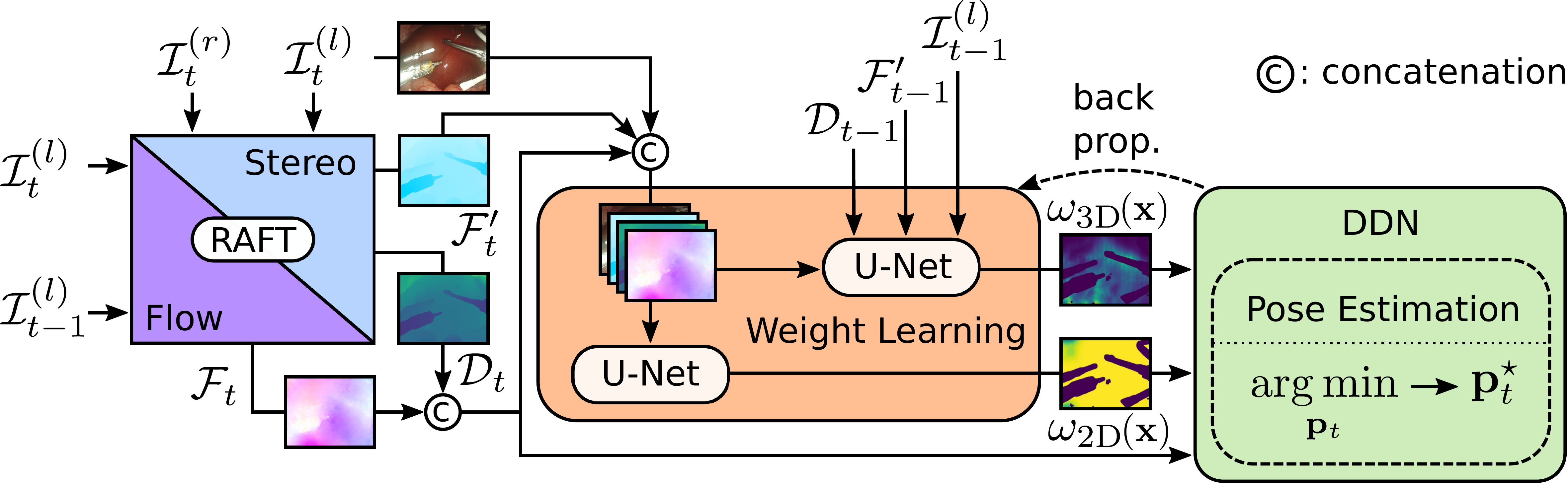}
    \caption{Overview of our proposed VO framework, which can be sub-divided into 3 parts (from left to right): (1) optical flow and depth are computed using RAFT~\cite{Teed2021}, (2) weight maps computed and used in (3) to estimate the camera pose $\mathbf{p}^\star_t$. Weight maps $(\omega_{\text{2D}}(\mathbf{x})$, $\omega_{\text{3D}}(\mathbf{x}))$ are learned via back propagation using the Deep Declarative Network (DDN)~\cite{Gould2021}.}
    \label{fig:method_overview}
\end{figure}

In the following, we present our \mbox{depth-based} pose estimation approach from an optimization perspective. We first derive our method in terms of context-specific adaptive weight maps in the pose estimation optimization and then show how to learn theset from data in an end-to-end way using DDNs to facilitate differentiation~\cite{Gould2021}. Our proposed method is illustrated in Fig.~\ref{fig:method_overview} and we provide a notation overview in~Tab.~\ref{tab:notations}.

For all images in an image sequence, we first employ RAFT~\cite{Teed2021} to establish correspondences between frames for both the stereo and temporal domain. This model allows disparity and optical flow estimation to be computed simultaneously, based on the observation that both share similar constraints on relative pixel displacements. From these estimates, we extract the horizontal component of the parallax flow $\mathcal{F}_t'$ at time $t$ as the stereo disparity to compute depth maps $\mathcal{D}_t$. As we would typically expect large vertical disparities in areas of low-texture or for de-calibrated stereo endoscopes, we use this parallax flow $\mathcal{F}_t'$ as input for the weight map estimation described in the following.

\subsection{Pose estimation}
In contrast to most existing VO methods, we estimate the camera motion exclusively based on a geometric loss function given that photometric consistency is entailed in optical flow estimation. We thus compute a 2D residual function based on a single depth map by,
\begin{align}
    r_{\text{2D}}(\mathbf{p}_t,\mathbf{x}) =\sqrt{\frac{1}{XY}} \left\lVert\pi_{\text{2D}}\Big(\exp(\mathbf{p}_t)\,\pi_{\text{3D}}\big(\mathcal{D}_{t},\mathbf{x}\big)\Big)-\big(\mathbf{x}+\mathcal{F}_t(\mathbf{x})\big)
    \right\rVert_2 \,\, ,
\end{align}
where $\pi_{\text{2D}}(\exp(\mathbf{p}_t)\,\pi_{\text{3D}}(\mathcal{D}_{t},\mathbf{x}))$ is the pixel location based on depth projection and the relative camera pose $p_t$ that aligns views $\mathcal{I}^{(l)}_t$ to $\mathcal{I}^{(l)}_{t-1}$. We scale these residuals by the image dimensions $X$ and $Y$ to make values independent of the image size. Note that we normalize depth maps by the maximum expected depth value, such that rotation and translation components of $\mathbf{p}_t$ have the same order of magnitude and thus equally contribute to the residuals, which is important for a well-conditioned optimization. Ideally, a projected pixel position coincides with the optical flow when the observed scene is rigid and when flow and depth maps are correct. 

\begin{table}[!t]
\begin{center}
\begin{minipage}[t]{\linewidth}
    \caption{Summary of used notation.}
    \label{tab:notations}
    \begin{tabular}{@{}ll@{}}
    Symbol & Description \\
    \toprule
    $t\in\mathbb{Z}$ & time frame index \\
    $\mathcal{I}^{(l)}_t
    \in\mathbb{R}^{X\times Y\times 3}$ & rectified left stereo image at time $t$ \\
    $\mathcal{D}_t\in\mathbb{R}^{X\times Y}$ & depth map w.r.t. to left image at time $t$ \\
    $\mathcal{F}_t\in\mathbb{R}^{X\times Y}$ & optical flow from $\mathcal{I}^{(l)}_t$ to $\mathcal{I}^{(l)}_{t-1}$ \\    
    $\mathcal{F}'_t\in\mathbb{R}^{X\times Y}$ & parallax flow displacement \if array\fi across stereo images \\
    $\mathbf{x}\in\mathbb{Z}^2$ & pixel index in 2D Cartesian coordinate system \\
    $\mathbf{p}_t\in\mathfrak{se}(3)\subset\mathbb{R}^6$ & relative pose from $t$ to $t-1$ in Lie algebra space \\
    $\mathbf{p}_t^\star\in\mathfrak{se}(3)\subset\mathbb{R}^6$ & relative pose solution in Lie algebra space \\
    $\exp(\mathbf{p}):\mathfrak{se}(3)\rightarrow\text{SE}(3)$ & matrix exponential \if mapping\fi from Lie algebra to Lie group \\
    $\pi_{\text{2D}}(\mathbf{v}):\mathbb{R}^4\rightarrow\mathbb{R}^2$ & projection of homogeneous 3D to 2D \if Cartesian\fi coordinates \\
    $\pi_{\text{3D}}(\mathcal{D}_t,\mathbf{x}):\mathbb{R}^{X\times Y}\times\mathbb{Z}^2\rightarrow\mathbb{R}^4$ & re-projection of 2D to 3D homogeneous coordinates \\
    $\mathcal{F}_t(\mathbf{x}):\mathbb{Z}^2\rightarrow\mathbb{R}^{2}$ & optical flow function across temporal domain\\
    $\omega_{\text{2D}}(\mathbf{x}):\mathbb{Z}^2\rightarrow\left[0, 1\right]$ & learned per-pixel weight for 2D residuals \\
    $\omega_{\text{3D}}(\mathbf{x}):\mathbb{Z}^2\rightarrow\left[0, 1\right]$ & learned per-pixel weight for 3D residuals \\
    $\lVert\cdot\rVert_n: \mathbb{R}^m\rightarrow\mathbb{R}^+$ & $\ell^n$ vector norm \\
    \botrule
    \end{tabular}
\end{minipage}
\end{center}
\end{table}

While minimizing $r_{\text{2D}}(\mathbf{p}_t,\mathbf{x})$ helps to reliably estimate the camera motion in rigid scenes, detection of deformations is most effective by looking at the displacement of points in 3D space. To address this need, we propose to leverage the depth map at $t-1$ and introduce a 3D residual function by,
\begin{align}
    r_{\text{3D}}(\mathbf{p}_t,\mathbf{x}) = \Big\lVert\exp(\mathbf{p}_t)\,\pi_{\text{3D}}\big(\mathcal{D}_t,\mathbf{x}\big)-\pi_{\text{3D}}\big(\mathcal{D}_{t-1},\mathbf{x}+\mathcal{F}_t(\mathbf{x})\big)\Big\rVert_2 \,\, ,
\end{align}
which measures the point-to-point alignment of the re-projected depth maps. As opposed to~\cite{Whelan2016}, we avoid using a point-to-plane distance as it is less constrained on planar surfaces such as organs (\eg, liver). While a known weakness of the point-to-point distance is its sensitivity to noise in regions with large perspectives, we mitigate this effect by combining 2D and 3D residuals. Intuitively, we expect $r_{\text{2D}}(\mathbf{p}_t,\mathbf{x})$ to be most accurate when camera motion is large and $r_{\text{3D}}(\mathbf{p}_t,\mathbf{x})$ when deformations are significant. Similar to~\cite{Teed2021}, we use bilinear sampling to warp point clouds from $\pi_{\text{3D}}\big(\mathcal{D}_{t-1},\mathbf{x}\big)$ to $\pi_{\text{3D}}\big(\mathcal{D}_{t-1},\mathbf{x}+\mathcal{F}_t(\mathbf{x})\big)$, using our optical flow estimates.

In contrast to conventional scalar-weighted sum of residuals, we propose to weigh each residual using a dedicated weight map that is inferred from the image data. The final residual is computed as,
\begin{align}
     r(\mathbf{p}_t,\mathbf{x})= \omega_{\text{2D}}(\mathbf{x})\,r_{\text{2D}}(\mathbf{p}_t,\mathbf{x}) + \omega_{\text{3D}}(\mathbf{x})\,r_{\text{3D}}(\mathbf{p}_t,\mathbf{x}) \,\, ,
    \label{eq:residuals_all}    
\end{align}
where $\omega_{\text{2D}}(\mathbf{x})$ and $\omega_{\text{3D}}(\mathbf{x})$ are the per-pixel weight maps for the 2D and 3D residuals, respectively. 

At its core, our hypothesis is that we can learn how the weight maps should combine the contributions of both $\omega_{\text{2D}}(\mathbf{x})$ and $\omega_{\text{3D}}(\mathbf{x})$ even in challenging situations where tissue deformations are taking place. That is, the role of  ($\omega_{\text{2D}}(\mathbf{x})$, $\omega_{\text{3D}}(\mathbf{x})$) is to (1) weigh relative focus based on the context of tissue deformations, (2) weigh reliable residual functions (2D vs 3D) given a motion pattern and (3) balance the scale of the loss. In Sec.~\ref{sec:learn_weights}, we detail how we learn a model to infer these weight maps. 

{\bf Optimization:} To compute the relative pose $\mathbf{p}_t^{\star} \in \mathfrak{se}(3)$, we then optimize, 
\begin{align}
\label{eq:optim}
    \mathbf{p}_t^{\star}=\underset{\mathbf{p}_t}{\operatorname{arg\,min}} \, \left\{\sum_{\mathbf{x}\in\boldsymbol{\Omega}} r(\mathbf{p}_t,\mathbf{x})^2\right\},
\end{align}
in a Non-Linear Least-Squares (NLLS) problem. Here, $\boldsymbol{\Omega}$ is a set containing all spatial image coordinates $\mathbf{x}$.
We choose to optimize the pose in the Lie algebra vector space $\mathfrak{se}(3)$ because this is a unique representation of the pose and has the same number of parameters as degrees of freedom. NLLS problems are typically solved iteratively using a second-order optimizer. To do this, we use the quasi-Newton method L-BFGS~\cite{LBGFS} due to its fast convergence properties and computational efficiency. Identical to~\cite{Parameshwara}, we simply chain relative camera poses to obtain the full trajectory.

\subsection{Learning the weight maps}
\label{sec:learn_weights}
In Eq.~\eqref{eq:residuals_all}, we propose to \textit{learn} residual weight maps $\omega_{\text{2D}}(\mathbf{x})$ and $\omega_{\text{3D}}(\mathbf{x})$, as determining these otherwise is not trivial. To this end, we train a separate encoder-decoder network, denoted by $g(\cdot)$, for each weight map. The input to these networks are all the elements used to compute residuals,
\begin{align}
    \omega_{\text{2D}}(\mathbf{x}) =& g\big(\mathbf{x},\mathcal{I}^{(l)}_t, \mathcal{D}_t, \mathcal{F}_t, \mathcal{F}'_t, \boldsymbol{\theta}_{\text{2D}}\big), \\
    \omega_{\text{3D}}(\mathbf{x}) =& g\big(\mathbf{x}, \mathcal{I}^{(l)}_t, \mathcal{D}_t, \mathcal{F}_t, \mathcal{F}'_t, \mathcal{I}^{(l)}_{t-1}, \mathcal{D}_{t-1},\mathcal{F}'_{t-1}, \boldsymbol{\theta}_{\text{3D}}\big) \, ,
\end{align}
where $\boldsymbol{\theta}_{\text{2D}}$ and $\boldsymbol{\theta}_{\text{3D}}$ are the network parameters that are learned at training time. For $g(\cdot)$, we employ a 3-layer UNet~\cite{unet_2015} with Sigmoid activation function to ensure outputs in $[0,1]$.

To train $g(\cdot)$, we aim to learn weight-maps that lead to improved pose estimation by minimizing the $\ell^1$ supervised training loss,
\begin{align}
    \mathcal{L}_{\text{train}}=\lVert \mathbf{p}^{\star}_{t}-\mathbf{p}^{(\text{gt})}_t\rVert_1,
\end{align}
where $\mathbf{p}^{(\text{gt})}_t$ is the ground-truth pose. Because the pose optimization in Eq.~\eqref{eq:optim} is not directly differentiable, we leverage a DDN~\cite{Gould2021} to enable end-to-end learning. This approach uses implicit differentiation of Eq.~\eqref{eq:optim} to compute the gradients of the weight map parameters $(\boldsymbol{\theta}_{\text{2D}}$, $\boldsymbol{\theta}_{\text{3D}})$ with respect to $\mathcal{L}_{\text{train}}$. Therefore, the only requirements are that (1) the function to be optimized $\sum_{\mathbf{x}\in\boldsymbol{\Omega}} r(\mathbf{p}_t,\mathbf{x})^2$ is twice differentiable and (2) we find a local or global minimum in the forward pass.

\section{Experiments}
\subsection{Datasets}
We evaluate our method on two separate stereo video datasets: one containing rigid MIS scenes and another containing non-rigid scenes:

\textbf{SCARED dataset~\cite{Allan2021}:} consists of 9 in-vivo porcine subjects with 4 sequences for each subject. The dataset contains a video stream captured using a da Vinci Xi surgical robot and camera forward kinematics. All sequences show rigid scenes without breathing motion or surgical instruments. We split the dataset into training (d2, d3, d6, d7) and testing sequences (d1, d8, and d9) where we exclude d4 and d5 due to bad camera calibrations.

\textbf{StereoMIS:} Additionally, we introduce a new in-vivo dataset also recorded using a da Vinci Xi surgical robot. Similarly to~\cite{Allan2021}, ground truth camera poses are generated from the endoscope forward kinematics and synchronized with the video feed. While we expect errors in the absolute camera pose due to accumulated errors in the forward kinematics, relative camera motions are expected to be accurate. It consists of 3 porcine (P1, P2, and P3) and 3 human subjects (H1, H2, and H3) with a total of 16 recorded sequences. We denote sequences with Px\_y where Px is the subject and y the sequence number. Sequence duration's range from 50 seconds to 30 minutes. They contain challenging scenes with breathing motions, tissue deformations, resections, bleeding and presence of smoke. We assign P1 and H1 to the training set and the rest is kept for testing. 

To provide a finer grained performance characterization of methods with this data, we extract from each video a number of short sequences that visually depict one of several possible settings:
\begin{itemize}
    \item \textit{breathing:} only depicts breathing deformations and contains no camera or tool motion,
    \item \textit{scanning:} includes camera motion in addition to breathing deformations,
    \item \textit{deforming:} comprises tissue deformations due breathing and manipulation or resection of tissue while the camera is static.
\end{itemize}
In practice, we select 88 different, non-overlapping, and 150-frames-long sequences from P2, P3, H2, H3 and assigned each to one of the above categories or {\it surgical scenarios} (see supplementary material for more information).

\subsection{Implementation details}
\subsubsection{Segmentation of surgical instruments}
For all experiments, we mask out surgical instrument pixels by setting corresponding residuals to 0. To do this, we use the DeepLabv3+ architecture~\cite{Chen_2018_ECCV} trained on the EndoVis2018 segmentation dataset~\cite{Allan2019} to generate instrument masks for each frame. Additionally, we mask out specularities, by means of maximum intensity detection, as they cause optical flow estimations to be ill-defined.

\subsubsection{Training and inference}
First, we classify all training frames from the SCARED and StereoMIS training sequences into "moving" and "static" based on the camera forward kinematics. We then randomly sample 4'000 frames from each sequence keeping a balance between moving and static frames. For each sampled frame, we generate a sample pair with an interval of 1 to 5 frames. We use the forward kinematics of the camera as the ground-truth pose change between two frames in a sample pair. Note that forward kinematics entail minor deviations that may propagate during training. We randomly assign $80\%$ of the sample pairs to the training set and $20\%$ to the validation set. 

For all experiments, we resize images to half resolution (512x640 pixels).
We use a batch size of 8 and the Adam optimizer with learning rate $10^{-5}$. We train for 200 epochs and perform early stopping on the validation loss. We implement our method using PyTorch, and train on a NVIDIA RTX3090 GPU. We reach 6.5 frames per second at test time. RAFT is trained on the FlyingThings3D dataset and we do not perform any fine-tuning.

\subsection{Metrics and baseline methods}
We use trajectory error metrics as defined in~\cite{ozyoruk2020}, namely the absolute trajectory error ATE-RMSE to evaluate the overall shape of the trajectory and the relative pose errors, RPE-trans and RPE-rot, to evaluate relative pose changes from frame to frame. The ATE-RMSE is sensitive to drift and depends on the length of the sequence, whereas the RPE measures the average performance for frame-to-frame pose estimation.

As no stereo SLAM method dedicated for MIS has open source code or is evaluated on a public dataset with trajectory ground-truth, we compare our method to two general state-of-the art rigid SLAM methods that contain loop closure and are based on the rigid scene assumption:
\begin{itemize}
    \item ORB-SLAM2~\cite{Mur-Artal2017}, a sparse SLAM leveraging bundle adjustment to compensate drift, 
    \item ElasticFusion~\cite{Whelan2016}, a dense SLAM and as such closer to our proposed method. 
\end{itemize}
In addition, we compare, our method to~\cite{Ruofeng2022} on the frames of the SCARED dataset for which they reported performances. For fair comparison, we use the same input depth maps for all methods.

\section{Results}

{\bf Surgical scenarios and ablation study:} 
Tab.~\ref{tab:scenarios} reports the performance of our approach on the StereoMIS surgical scenarios. To show the importance of learning the weight maps we perform an ablation study where we evaluate the impact of (1) constant weights, denoted by \textit{ours (w/o weight)}, where $\omega_{\text{2D}}(\mathbf{x})=\omega_{\text{3D}}(\mathbf{x})=1$ for each $\mathbf{x}$; (2) our method with only 2D-residuals, denoted by \textit{ours (only 2D)}; and (3) using only 3D-residuals, denoted by \textit{ours (only 3D)}.
\begin{table}[!t]
    \centering
    \caption{The ATE-RMSE (mean$\pm$std mm) for the different scenarios from the StereoMIS dataset with average over sequences (micro avg.) and scenarios (macro avg.).}
    \addtolength{\tabcolsep}{-0.3em} 
    \begin{tabular}{@{}l|ccc|cc@{}}
        scenario & \textit{breathing} & \textit{scanning} & \textit{deforming} & micro avg. & macro avg.\\
         \# sequences &17&60&9 & &\\
        camera motion   &  &\checkmark & &  \\
        breathing  &\checkmark  &\checkmark  &\checkmark  & &   \\
        tool interactions  &  &  &\checkmark  & & \\
        
    \midrule
       ORB-SLAM2~\cite{Mur-Artal2017}  &  $2.35\pm 1.81$& $3.26 \pm 1.65$ & $4.29 \pm 2.30$ & $3.19 \pm 1.81$ & $3.30 \pm 0.97$\\
       ElasticFusion~\cite{Whelan2016} &  $1.94 \pm 0.93$&$4.04 \pm 3.46$& $6.47 \pm 8.64$  &$3.88 \pm 4.12$ &$4.15 \pm 2.27$ \\
       \midrule
       \bf{ours} (w/o weight) & $1.65\pm 0.97$ &  $3.01 \pm 1.60$& $4.67 \pm 2.13$& $2.91 \pm 1.74$&   $3.11 \pm 1.51$\\
       \bf{ours} (only 2D)  & $1.15\pm 0.72$ &  $3.01 \pm 1.66$& $2.83 \pm 1.41$ & $2.62 \pm 1.66$& $2.33 \pm 1.03$  \\
       \bf{ours} (only 3D)  & $\mathbf{0.78\pm 2.03}$ &  $7.02 \pm 5.86$& $2.72 \pm 1.90$& $5.34 \pm 5.64$&  $3.51 \pm 3.20$  \\
       \bf{ours} (2D \& 3D) & $1.01 \pm 0.59$ &  $\mathbf{2.89\pm 2.33}$& $\mathbf{2.23 \pm 1.07}$ &$\mathbf{2.45 \pm 2.12}$ & $\mathbf{2.04 \pm 0.95}$ 
    \label{tab:scenarios}
    \end{tabular}
\end{table}

Our proposed method outperforms the baselines in all scenarios. Improvements in \textit{breathing} and \textit{scanning} are partly due to correct identification of errors in the optical flow and depth estimation as well as optimal balancing of the 2D and 3D residuals. Indeed, exploiting the complementary properties of 2D and 3D residuals improves the average performance. The fact, that ours (only 3D) outperforms ours (only 2D) in \textit{breathing} and \textit{deforming} supports our intuition that it is easier to learn tissue deformations from the 3D residuals. Contrarily, in \textit{scanning} where the optical flow is dominated by the camera motion, the 2D residuals lead to a more accurate pose estimation. 

In general, it is not possible to detect or completely compensate the breathing motion on a frame-to-frame basis in our proposed optimization scheme as we cannot completely disambiguate the camera and tissue motion. However, the method learns which regions are more affected by breathing deformations and consequently assigns a smaller weight to those regions. 

\begin{figure}[!t]%
\centering
\includegraphics[width=\textwidth]{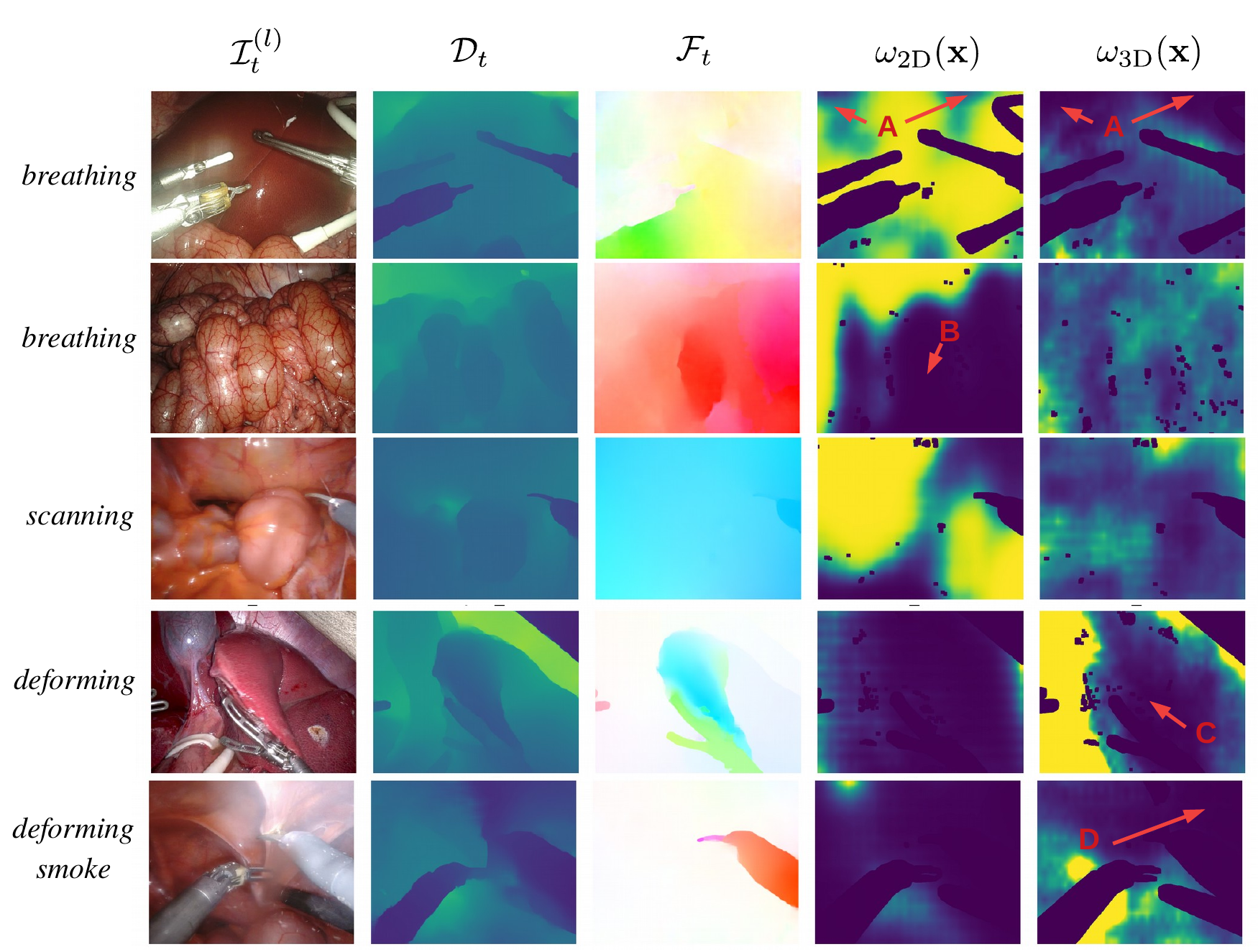}
\caption{Exemplary results for 5 different scenarios. Surgical instruments and specularities are masked out. 
From left to right: left image, its depth map, its optical flow displacement as well as its weights $\omega_{\text{2D}}(\mathbf{x})$ and $\omega_{\text{3D}}(\mathbf{x})$. Weight maps are normalized (lowest value in dark blue and highest value in yellow). Best viewed in color.
}
\label{fig:weight_maps}
\end{figure}

We note that the weight maps in Fig.~\ref{fig:weight_maps} (see~\textit{breathing} rows) support our claims that the weight maps have low values in the dark regions (A) where we expect the optical flow to be inaccurate and where tissue is moving most (B). The \textit{scanning} example also illustrates that the weight maps have a different response depending on the motion pattern and deformation.
Note that the presence of surgical instruments has no influence on the weight maps in \textit{scanning}, as no tissue interaction takes place. As expected, the largest improvements can be seen in the \textit{deforming} scenario. Inspecting the two last rows in~Fig.~\ref{fig:weight_maps} reveals that regions where the instruments deforms tissue (C) are correctly ignored in the pose estimation. Similarly, the region occluded by smoke (D) has low values in the weight maps. Additionally, we observe that $\omega_{\text{2D}}$ usually has 100 times larger magnitude than $\omega_{\text{3D}}$ compensating for the different scale of the 2D and 3D residuals. 

\begin{figure}[!b]%
\centering
\includegraphics[width=0.9\textwidth,trim={0 0.25cm 0 0.2cm},clip]{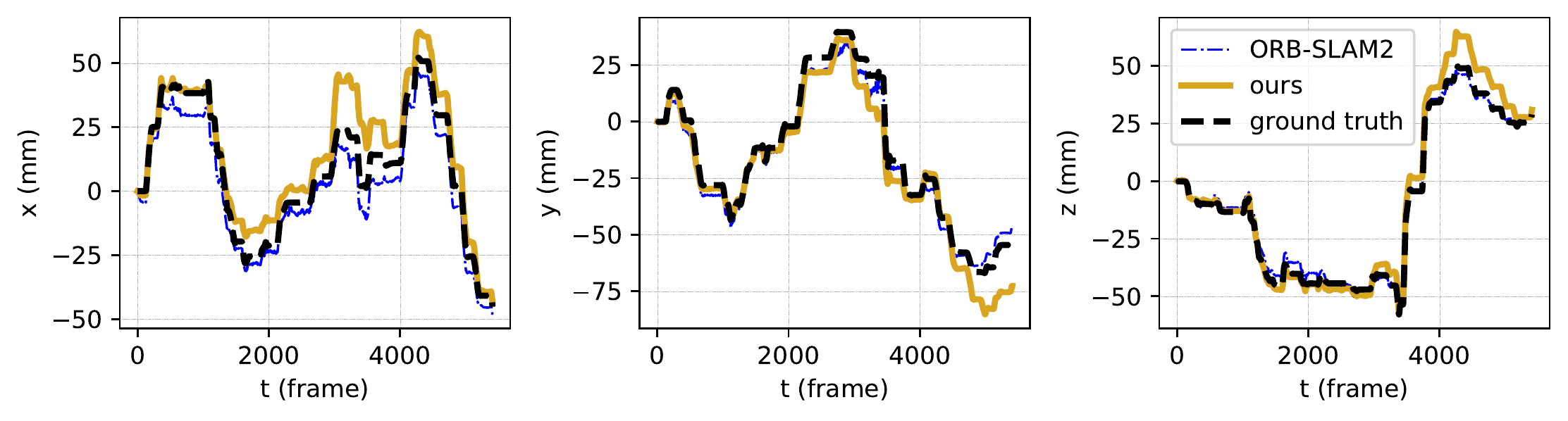}
\includegraphics[width=0.9\textwidth,trim={0 0.25cm 0 0.2cm},clip]{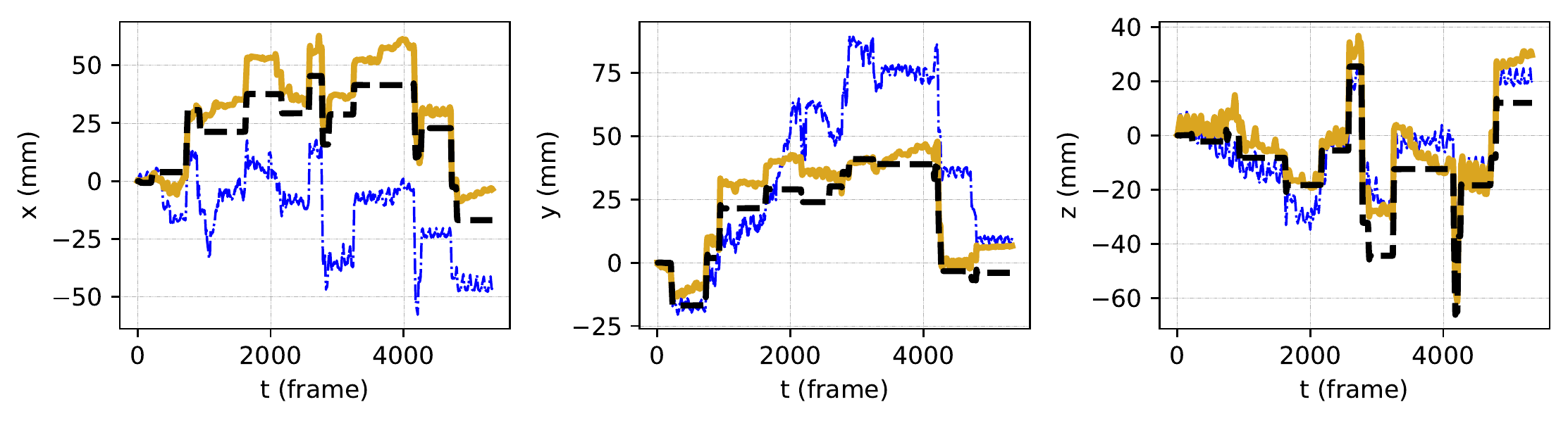}
\caption{Two example trajectories of the \textit{StereoMIS} test set. \textbf{Top: }~trajectory of P2\_7. \textbf{Bottom: }~trajectory of P3\_0. }\label{fig:traj}
\end{figure}

{\bf Results on full test StereoMIS sequences: }
Tab.~\ref{tab:test} shows the pose estimation performance on the complete sequences in the StereoMIS testset. As the sequences are much longer than in the scenario experiment, accumulation of drift results in a large ATE-RMSE for all methods. Even though our frame-to-frame approach does not include any bundle-adjustment or regularization over time, it still has the lowest ATE-RMSE on average. The reason for this good performance is reflected in the relative metrics RPE-trans and RPE-rot, where our method outperforms all others by almost a factor of three and five, respectively. Our method robustly estimates the pose in challenging situations whereas ORB-SLAM2 fails in two sequences (H2\_0, P2\_5). Fig.~\ref{fig:traj} shows two example trajectories. P2\_7 does not include any tool-tissue interactions and consists of smooth camera motions. Its trajectory illustrates the drift of our method which results in an ATE-RMSE of $9.28$ versus $3.76$mm for ORB-SLAM2. On the other hand, P3\_0 consists of strong tissue deformations and abrupt camera movements. Despite visible drift, we can see that our method is able to follow the abrupt movements. The small scale oscillations in the trajectories are due to breathing motion. The trajectories of all test sequences and evaluation results excluding frames where the SLAM methods fail can be found in the supplementary materials.
\newcommand{\STAB}[1]{\begin{tabular}{@{}c@{}}#1\end{tabular}}
\begin{table}[t!]
    \centering
    \caption{Pose estimation results on full StereoMIS test sequences for ORB-SLAM2~\cite{Mur-Artal2017}, ElasticFusion~\cite{Whelan2016}, and \textbf{ours}. Metrics are reported in (mean$\pm$std) when applicable.}
    \label{tab:test}
    \addtolength{\tabcolsep}{-0.3em} 
    \begin{tabular}{@{}llllll@{}}
    \toprule
         & H2 & H3 & P2 & P3 & macro avg.  \\
    \midrule
    
    \multicolumn{6}{c}{ATE-RMSE (mm)}\\
        ORB-SLAM2~\cite{Mur-Artal2017} & $18.0$ & $\mathbf{9.1}$& $14.0$ &  $21.4 $ & $15.6 \pm 5.3$ \\ 
        ElasticFusion~\cite{Whelan2016} & $30.8 $ & $72.1$ & $33.6$  & $37.7$& $43.6 \pm 19.3$ \\
        
        \bf{ours} & $\mathbf{10.9}$ & $21.2$& $\mathbf{13.8}$  & $\mathbf{8.8}$& $\mathbf{13.7 \pm 5.4}$  \\
    \midrule
        
        \multicolumn{6}{c}{RPE-trans (mm)}\\ 
        ORB-SLAM2~\cite{Mur-Artal2017} & $0.20 \pm 0.43$& $0.24 \pm 0.25$ & $0.35 \pm 0.46$  & $0.54 \pm 0.47$ & $0.33 \pm 0.13$ \\ 
        ElasticFusion~\cite{Whelan2016} & $0.87 \pm 1.11$ & $0.56 \pm 1.03$& $0.81 \pm 1.11$  & $0.71 \pm 0.79$& $0.74 \pm 0.12$ \\
        
       \bf{ours}  & $\mathbf{0.10 \pm 0.27}$& $\mathbf{0.10 \pm 0.18}$ & $\mathbf{0.16 \pm 0.32}$  & $\mathbf{0.19 \pm 0.31}$& $\mathbf{0.14 \pm 0.04}$  \\
    \midrule
        
        \multicolumn{6}{c}{RPE-rot (deg)}\\
        ORB-SLAM2~\cite{Mur-Artal2017} & $0.16 \pm 0.36$& $0.16 \pm 0.22$ & $0.19 \pm 0.24$  & $0.28 \pm 0.27$ & $0.20\pm 0.05$ \\ 
        ElasticFusion~\cite{Whelan2016} & $0.73 \pm 1.06$& $0.41 \pm 0.96$ & $0.50 \pm 1.11$  & $0.38 \pm 0.40$& $0.51 \pm 0.14$ \\
        
        \bf{ours}  & $\mathbf{0.04 \pm 0.20}$& $\mathbf{0.04 \pm 0.13}$ & $\mathbf{0.07 \pm 0.14}$  & $\mathbf{0.05 \pm 0.10}$& $\mathbf{0.05 \pm 0.01}$  \\
    \botrule
    \end{tabular}
\end{table}

{\bf Results on SCARED dataset: }
Wei~\textit{et al.} reported ATE-RMSE results for rigid surgical scenes of the SCARED dataset in a frame-to-model approach~\cite{Ruofeng2022}. For the sake of fair comparison, we extend our method to SLAM by accommodating a surfel map model denoted by \textit{ours (frame2model)}, which is equivalent to that used in~\cite{Ruofeng2022}. Specifically, we replace input images $\mathcal{I}^{(l)}_{t-1}, \mathcal{I}^{(r)}_{t-1}$ by rendered images from the surfel map. Note, we can only adopt this approach for the SCARED dataset, as the surfel map model assumes scene rigidity. Results are provided in Tab.~\ref{tab:scared}.

\begin{table}
    \centering
    \caption{The ATE-RMSE in mm for SCARED sequences reported by~\cite{Ruofeng2022} and micro average over all SCARED test sequences (SCARED avg.) using surfel maps. }
    \begin{tabular}{@{}llllll|l@{}}
    \toprule
         & d1\_k2 & d8\_k1 & d9\_k1 & d9\_k3  & avg. & SCARED avg. \\
    \midrule
        ORB-SLAM2~\cite{Mur-Artal2017} & $0.91$ & $2.97$ & $4.33$ & $3.79$ & $3.00$ & $2.34 \pm 1.24$\\ 
        ElasticFusion~\cite{Whelan2016} & $1.02$ & $3.62$ & $4.30$ & $3.36$ & $3.08$ & $2.91 \pm 1.77$\\ 
        Wei~\textit{et al.}~\cite{Ruofeng2022} & $0.74$ & $2.47$ & $4.07$ & $1.54$ & $2.21$ & -\\ 
        \bf{ours} (frame2model) & $\mathbf{0.37}$ & $\mathbf{2.08}$ & $\mathbf{2.04}$ & $\mathbf{0.84}$ & $\mathbf{1.33}$ & $\mathbf{1.38 \pm 0.93}$\\
    \botrule
    \label{tab:scared}
    \end{tabular}
\end{table}
\section{Conclusion}
\label{sec:conclusion}

We proposed a visual odometry method for robust pose estimation in the challenging context of endoscopic surgeries.
To do so, we learn adaptive weight maps for two geometrical residuals to leverage pose estimation performance on common surgical scenes including breathing motion and tissue deformations.
Thanks to a performance analysis in common scenarios, we observed the complementary action of the 2D/3D residuals and the strong contribution of their specific weighting at pixel level.
This results in better performances compared to state-of-the-art methods, on average and in the most challenging cases.
We believe that our contributions are beneficial for some SLAM components, \textit{e.g.} map building, and therefore downstream applications in MIS.
Future work will focus on drift and breathing compensation.

\backmatter

\bmhead{Supplementary information}
Appendix A: details on \textit{StereoMIS}. Appendix B: trajectories of \textit{StereoMIS} test set. Appendix C: Results on full \textit{StereoMIS}
sequences. Appendix D: trajectories of \textit{SCARED} test set.

\section*{Declarations}

\begin{itemize}
\item Funding: This work was supported by InnoSuisse grant \# 50204.1 IP-LS. 
\item Competing interests: Authors declare that they have no conflict of interest.
\item Ethics approval:
All applicable international, national, and institutional guidelines for the care and use of animals were followed. All procedures performed in studies involving animals were in accordance with the ethical standards of the institution at which the studies were conducted.
\item Consent: N/A
\item Availability of data, materials, and code: \\ \textit{StereoMIS} porcine data - \url{https://doi.org/10.5281/zenodo.7727692}. \\Code and models - \url{https://github.com/aimi-lab/robust-pose-estimator}
\end{itemize}

\bibliography{ref}

\end{document}